\documentclass{ecai} 

\usepackage{latexsym}
\usepackage{amssymb}
\usepackage{amsmath}
\usepackage{amsthm}
\usepackage{booktabs}
\usepackage{enumitem}
\usepackage{graphicx}
\usepackage{color}

\usepackage{mathtools}
\usepackage{physics}

\usepackage{caption}
\usepackage{subcaption}
\usepackage{xcolor}
\usepackage{dsfont}
\usepackage{hyperref}

\newcommand{\BibTeX}{B\kern-.05em{\sc i\kern-.025em b}\kern-.08em\TeX}

\definecolor{1st}{HTML}{D81B60}
\definecolor{2nd}{HTML}{1E88E5}
\definecolor{3rd}{HTML}{FFC107}
\definecolor{4th}{HTML}{004D40}
\usepackage{verbatim}

\begin{document}


\begin{frontmatter}


\paperid{8511} 


\title{UniMLR: Modeling Implicit Class Significance for Multi-Label Ranking}


\author[A]{\fnms{V. Bugra}~\snm{Yesilkaynak}\thanks{Corresponding Author. Email: bugra.yesilkaynak@tum.de.}\footnote{Equal contribution.}\footnote{Work done at Istanbul Technical University.}}
\author[A]{\fnms{Emine}~\snm{Dari}\footnotemark[1]\footnotemark[2]}
\author[C]{\fnms{Alican}~\snm{Mertan}}
\author[B, D]{\fnms{Gozde}~\snm{Unal}}

\address[A]{Technical University of Munich}
\address[B]{Istanbul Technical University}
\address[C]{University of Vermont}
\address[D]{New York University}


\begin{abstract}
Existing multi-label ranking (MLR) frameworks only exploit information deduced from the bipartition of labels into positive and negative sets. Therefore, they do not benefit from ranking among positive labels, which is the novel MLR approach we introduce in this paper. We propose UniMLR, a new MLR paradigm that models implicit class relevance/significance values as probability distributions using the ranking among positive labels, rather than treating them as equally important. This approach unifies ranking and classification tasks associated with MLR. Additionally, we address the challenges of scarcity and annotation bias in MLR datasets by introducing eight synthetic datasets (Ranked MNISTs) generated with varying significance-determining factors, providing an enriched and controllable experimental environment. We statistically demonstrate that our method accurately learns a representation of the positive rank order, which is consistent with the ground truth and proportional to the underlying significance values. Finally, we conduct comprehensive empirical experiments on both real-world and synthetic datasets, demonstrating the value of our proposed framework. Code is available at \href{https://github.com/MrGranddy/UniMLR}{https://github.com/MrGranddy/UniMLR}.
\end{abstract}

\end{frontmatter}


\section{Introduction}
Multi-label ranking (MLR) \cite{Zhou2014ATO,Dery2021MultilabelRM,InconsistentRankers} is a supervised learning problem that focuses on both identifying the relevant labels for a given instance and determining the order of these labels based on their relevance. It can be considered as a combination of multi-label classification, where multiple labels can be assigned to an instance, and label ranking, where the labels are sorted according to their relevance. More formally, given a set of labels $Y$, multi-label classification (MLC) bipartites the set into two and associates the instances with the relevant label set $\mathcal{Y} \subseteq Y$. On the other hand, label ranking aims to map instances to a total order over $Y$. With the objectives of these sub-problems combined, multi-label ranking is applicable in scenarios where the expected output is a ranked subset of all possible labels \cite{10.5555/1567016.1567123, Brinker2007CaseBasedMR}. Practical applications of multi-label ranking span diverse domains. For example, in recommendation, items can belong to multiple categories that must be ranked by relevance \cite{li2023recommendersystems}; in computer vision, images may have multiple labels whose importance varies by context \cite{DBLP:journals/corr/LiSL17}; in bioinformatics, genes can have multiple functions that exist in a hierarchy \cite{KDDGene}.

However, MLR datasets, which include additional information on the order of relevant labels, are less common than MLC datasets, where all relevant labels are equally important. Additionally, MLR datasets may suffer from subjective or biased label rankings \cite{InconsistentRankers}. Consequently, studies that evaluate the performance of proposed approaches based on the ranking order of predicted positive classes are also less common. Instead, MLR is often used to address MLC problems, such as in object recognition \cite{Bucak2009EfficientMR}, image classification \cite{DBLP:journals/corr/LiSL17}, and online algorithms \cite{pmlr-v84-jung18a}, where the ranking of predictions is not a primary performance measure in the end. Furthermore, we notice that the existing approaches, both focusing on MLR as a learning task or utilizing MLR as an auxiliary method, rely on recovering true label ranks from limited information derived from binary labels (positive or negative), rather than utilizing the rank orders of positive labels.

Given the limitations, the motivation for our work is to study MLR with a novel approach that exploits the order between positive labels, instead of assuming the positive and negative labels of equal importance within their sets. We highlight our contributions and focus of our study as follows:
\begin{itemize}

    \item  In the field of MLR, we establish the paradigm of exploiting the ranking information between positive labels of an instance, aiming to extract significance values of labels that determine their rank, which may not be feasible to obtain numerically during the labeling process.
    
    \item We model the MLR problem as a distribution learning problem based on a probabilistic approach that unifies the bi-partition and ranking of the labels in the same space. This way, we incorporate pairs of positive labels into the optimization in addition to positive and negative label pairs, to reveal a preference relation over a varying number of positive labels.
    
    \item We introduce ranked image datasets generated under different setups with varying importance factors that determine the ranks, which create a controllable environment to test new approaches while facilitating unambiguous interpretations of their performances.
    
    \item We compare our novel framework with different related methods, empirically explore and interpret the outcomes, thus defining clear baselines for the MLR problem.
    
\end{itemize}

\section{Related Work}
In the literature, MLR has been used to classify and rank scene categories in natural scenery images or emotions from facial expressions \cite{InconsistentRankers} as real-world use cases, as well as being applied to solve classification problems \cite{Bucak2009EfficientMR,DBLP:journals/corr/LiSL17,PositivePairwiseCorrelations,DembczynskiConsistent}, to learn from incompletely or inconsistently labeled data \cite{InconsistentRankers,PositiveUnlabeled}, and has been studied from perspectives of consistency and generalization by \cite{RethinkingWu}. 
In the following subsections, we review related work on the two sub-problems associated with MLR. We do not compare UniMLR with works in label distribution learning (LDL) \cite{XinGeng2016}, as they aim to predict a real-valued overall distribution and therefore work with real-valued ground truth annotations, unlike integer ranks as in MLR, thus lying beyond the scope of fair comparison. Although certain studies may seem similar,  such as predicting label distributions from MLR-annotations \cite{10198360, NEURIPS2022_efc549c2}, the learning task differs from MLR, there is no partitioning of the labels as relevant or irrelevant as in the multi-label learning field and the performance is mostly measured by distance-based metrics.
Some works have explored the ranking order between positive labels \cite{8892640,casebased}, but were short of employing this information to guide the learning process as our work does with positive label pairs.
\paragraph{Label ranking.}
There are a variety of approaches to solve label ranking, where extensive overview is provided by \cite{Zhou2014ATO,Vembu2010LabelRA}. Related methods can be roughly divided into three:  pointwise \cite{pointChen,pointSigur,pointToderici}, pairwise \cite{hullermeier_2008,Zhou2014ATO,DBLP:journals/corr/LiSL17,mertan_new_2020} and listwise \cite{listwiseCao,listwiseXia}, where our work mainly lies in the area of pairwise methods. A pairwise method transforming the problem into binary classification was introduced in the Constraint Classification framework (CC) \cite{HarPeled2002ConstraintCF}. CC builds constraints according to the preference relation between labels, where the relation $\lambda_i > \lambda_j$, denoting that label $\lambda_i$ precedes $\lambda_j$ in relevance to the instance, constructs the positive constraint $f_i(x) - f_j(x)>0$ and the negative constraint $ f_j(x) - f_i(x)<0$. Then, both constraints are used as training samples for the single classifier. More efficiently, Ranking by Pairwise Comparison (RPC) \cite{hullermeier_2008,Frnkranz2010PreferenceLA} transforms the problem into training a binary classifier for each label pair, producing a number of $K(K-1)/2$ models which is half of the number of constraints of CC, where $K$ is the number of classes. The final ranking is determined by the votes of each model for the given pair. More recently, the Log-Sum-Exp-Pairwise (LSEP) loss function introduced in  \cite{DBLP:journals/corr/LiSL17}
improves the previous approaches using hinge loss \cite{Gong2014DeepCR,Weston2011WSABIESU} by learning the pairwise comparisons in a smooth and easier way to optimize, by pairing one positive and one negative label in ground truth then wrapping BP-MLL (Backpropagation for Multi-Label Learning) \cite{ZhangBPMLL} in a logarithmic function with a bias term, with the objective of enforcing the positive labels to be in higher ranks and negative labels to be in lower ranks. However, LSEP involves an obligatory ordering of the whole set of labels to be followed by a thresholding that determines which labels are to be discarded, while our UniMLR introduces a natural label selection and ranking process.

\paragraph{Label classification.}
Classification algorithms can be distinguished by how many labels they assign to each instance. In single-label classification, exactly one label is predicted, whereas in multi-label classification, an instance can be associated with multiple labels. We refer to this number of predicted labels as the label count. Our work relates to the latter, namely multi-label classification. In multi-label setting, different strategies exist to determine the label count. A simple approach is to set a fixed label count, e.g., always choosing the top-k scoring labels, or applying a fixed threshold on the prediction confidence. However, such heuristics are often impractical, as they ignore problem-specific context. More flexible methods instead allow the label count to vary. For example, varying boundaries can be learned by training with adaptive thresholds or label counts \cite{DBLP:journals/corr/LiSL17}, or incorporated into the label ranking process \cite{10.5555/1567016.1567123} by introducing virtual “split” labels that define cut-off points, which are inserted  in the label set before the ranking process. In contrast, our UniMLR implicitly introduces a zero-point, i.e. an inherent threshold, to perform binary classification of labels into positives and negatives in the same space that we perform ranking, thus combining both tasks in a unified model.

\section{Problem and Notation}

\subsection{Dataset definition}
\label{sec:dataset_notation}
We start with a dataset of $N$ examples, $\{(\textbf{\textit{x}}^{(i)}, R^{(i)})\}^N_{i=1}$ where $\textbf{\textit{x}}^{(i)} \in \mathbb{R}^d$ is a real-valued sample from the input distribution, and $R^{(i)} = \{(y_j, r^{(i)}_j)\}_{j=1}^K$ is the label, where $y_j \in Y$ is a symbolic class representation for an entity that can be semantically present in $\textbf{\textit{x}}^{(i)}$, $Y = \{y_1, y_2, ..., y_K\}$ is the set of $K$ possible classes, and $r^{(i)}_j \in \mathbb{N}$ is the rank of the associated class $y_j $ for $\textbf{\textit{x}}^{(i)}$. It is the case that prior work mostly uses standard multi-label classification datasets defined as $\{(\textbf{\textit{x}}^{(i)}, \mathcal{Y}^{(i)})\}^N_{i=1}$ where $\mathcal{Y}^{(i)} \subseteq Y$ is the set of classes semantically present in $\textbf{\textit{x}}^{(i)}$ and called the \textit{positive} classes, while the rest of the classes are called \textit{negative} classes. This kind of dataset only provides a ranking information between negative and positive classes where positives have a higher rank than negatives, and can be seen as a special case of the former definition where $R^{(i)} = \{(y_u, 1) | y_u \in \mathcal{Y}^{(i)}\} \cup \{(y_v, 0) | y_v \notin \mathcal{Y}^{(i)}\} $. In our definition, a negative class will always have rank 0. Throughout our work, our findings are under the fair assumptions: \textbf{(i)} $\textbf{\textit{x}}$ are identically and independently distributed (i.i.d.) in all datasets. \textbf{(ii)} All label-significance value pairs $(y_j, s_j^{(i)})$ are conditionally independent given an input $\vb*{x^{(i)}}$ where a significance value is the underlying importance that describes the ranks.

\subsection{Problem definition}

We construct the \textit{multi-label ranking} problem, assuming that the two sub-problems, the multi-label classification and label ranking, are independent, i.e., $\mathcal{Y}^{(i)}$ (positive label set) and $\mathcal{B}^{(i)}$ (bucket order) are independent given $\vb*{x}^{(i)}$. Thus, we formally define the following likelihood optimization problem, given an instance $\vb*{x}^{(i)}$:
\begin{equation}\label{eqn:mlr}
    \max_{\theta,\phi} P_\mathcal{Y}(\mathcal{Y}^{(i)}|\textbf{\textit{x}}^{(i)};\theta)P_\mathcal{B}(\mathcal{B}^{(i)}|\textbf{\textit{x}}^{(i)};\phi).
\end{equation}
Here $\mathcal{Y}^{(i)}$ is the set of \textit{positive} classes for an instance $\textbf{\textit{x}}^{(i)}$, $P_\mathcal{Y}$ is the parameterized family of probability mass functions of possible positive class sets, parameterized by $\theta$, and conditioned by $\textbf{\textit{x}}^{(i)}$. 
While $\mathcal{B}^{(i)}$ is a \textit{bucket order} \cite{10.1145/1055558.1055568}, which is a class of partial orders allowing \textit{ties}. A partial order is a reflexive, antisymmetric, and transitive binary relation on a set of items, in our case $Y$. $\mathcal{B}^{(i)}$ intuitively partitions labels $y \in Y$ into mutually exclusive bucket of ranks $\langle \mathcal{M}_1^{(i)}, ..., \mathcal{M}_{b^{(i)}}^{(i)} \rangle$ where $b^{(i)}$ is the number of buckets. If two items $y_u, y_v \in \mathcal{M}_k^{(i)}$ are the member of the same bucket then we say they are in tie, meaning they can not be distinguished ordinally and they virtually have the same rank. Formally, for a bucket $\mathcal{M}_k^{(i)}$, ~$y_u, y_v \in \mathcal{M}_k^{(i)} \iff (y_u, y_v) \notin \mathcal{B}^{(i)} \wedge (y_v, y_u) \notin \mathcal{B}^{(i)}$. Here it should be noted that $\mathcal{M}_k^{(i)} \subseteq Y$ is a set and is only introduced to better visualize the bucket orders, $\mathcal{B}^{(i)}$ is just a relation and is enough to define a bucket order. On the other hand, different buckets for an instance $x^{(i)}$ have total ordinal relationship between them, such that: for any two distinct buckets $\mathcal{M}_k^{(i)}, \mathcal{M}_l^{(i)}$, $y_u \in \mathcal{M}_k^{(i)}, y_v \in \mathcal{M}_l^{(i)}$, $(y_u, y_v) \in \mathcal{B}^{(i)} \iff k > l$, i.e. $r_u^{(i)} \geq r_v^{(i)} \iff k > l$, where $r_u^{(i)}$ and $r_v^{(i)}$ are corresponding ranks of $y_u$ and $y_v$. $P_{\mathcal{B}}$ is the parameterized family of probability mass functions of such bucket orders parameterized by $\phi$, conditioned by $\textbf{\textit{x}}^{(i)}$.

The optimization problem in hand can be seen as the joint optimization of two distinct problems, namely: multi-label classification and label ranking.
Both of the terms can be divided into practicable sub-problems.

We can simplify the first likelihood  $P_{\vb{y}}(\vb{y}|\textbf{\textit{x}}^{(i)};\theta)$ by defining the random variable $\vb{y} \in \{0, 1\}^K$ via the random vector:
\begin{align*}
    &\vb{y}_c =
    \begin{cases} 
      1 & y_c \in \mathcal{Y} \\
      0 & y_c \notin \mathcal{Y}
   \end{cases}, &c \in \{1, ..., K\}.
\end{align*}
Here $\mathcal{Y}$ is any random variable distributed by $P_\mathcal{Y}(\mathcal{Y}|\textbf{\textit{x}}^{(i)};\theta)$, then we can model the sub-problem with $K$ binary classification models using Bernoulli distribution:

\begin{equation}\label{eqn:bernoulli_part}
    \begin{split}
        &\max_\theta P_\mathcal{Y}(\mathcal{Y}^{(i)}|\textbf{\textit{x}}^{(i)};\theta) 
        = \max_\theta \prod_{c=1}^K B(\vb{y}_c, \textbf{\textit{x}}^{(i)}, \theta), \\
        \text{where }& B(\vb{y}_c, \textbf{\textit{x}}^{(i)}, \theta) = 
        \begin{cases} 
            P(\vb{y}_c=1|\textbf{\textit{x}}^{(i)};\theta) & \text{if } \vb{y}_c \in \mathcal{Y}^{(i)}, \\
            P(\vb{y}_c=0|\textbf{\textit{x}}^{(i)};\theta) & \text{if } \vb{y}_c \notin \mathcal{Y}^{(i)}.
        \end{cases}
    \end{split}
\end{equation}

We parameterize $P_{\vb{r}}(\vb{r}|\textit{\textbf{x}}^{(i)};\phi)$ by $\phi$, and re-write $P(\mathcal{B})$ in a likelihood maximization:

\begin{equation}\label{eqn:bucket_prob}
\begin{split}
         &\max_\phi P_\mathcal{B}(\mathcal{B}^{(i)}|\textbf{\textit{x}}^{(i)};\phi)  = \max_\phi \smashoperator{\prod_{(y_u, y_v) \in \mathcal{B}^{(i)}}} P(\vb{r}_u \geq \vb{r}_v|\textit{\textbf{x}}^{(i)};\phi).
\end{split}
\end{equation}

As a result, the multi-label ranking problem as defined in Equation~\eqref{eqn:mlr}, using Equation~\eqref{eqn:bernoulli_part} and Equation~\eqref{eqn:bucket_prob} can be re-written as:
\begin{equation}\label{eqn:simple_mlr}
    \max_{\theta,\phi} \left[ \prod_{c=1}^K B(\vb{y}_c, \textbf{\textit{x}}^{(i)}, \theta) \smashoperator{\prod_{(y_u, y_v) \in \mathcal{B}^{(i)}}} P(\vb{r}_u \geq \vb{r}_v|\textit{\textbf{x}}^{(i)};\phi) \right].
\end{equation}

The refined optimization problem in Equation~\eqref{eqn:simple_mlr} sets up the basis for our proposed unified multi-label ranking method. It establishes a probabilistic foundation for the multi-label ranking problem, which we further develop with a Gaussian probability model next.

\section{UniMLR}

\subsection{Core idea}
Multi-label ranking problem can be viewed as both classifying and ranking the classes in a given instance, meaning that we are not only interested in assigning correct classes to an instance, but we also want to measure how relevant these classes are for the given instance.

\textbf{Significance values.} Consider each $y_j \in \mathcal{Y}^{(i)}$ associated with a given input $\vb*{x}^{(i)}$. We attribute to each $y_j$ a corresponding significance value $s_j^{(i)} \in \mathbb{R}$. These significance values are initially posited to adhere to a predefined probability distribution, with the choice of this distribution being flexible. Our model remains agnostic to this distribution's specific nature, allowing for adaptable application across various contexts. 

In this framework, we choose to model the significance values using Gaussian distributions. We selected Gaussian distributions for their simplicity, symmetry, and widespread use in related areas such as LDL. Therefore, we define each significance value as $s_j^{(i)} \sim \mathcal{N}(\mu_j^{(i)}, \sigma_j^{2(i)})$, where $\mathcal{N}(\cdot,\cdot)$ denotes a Gaussian distribution with mean $\mu_j^{(i)}$ and variance $\sigma_j^{2(i)}$, corresponding to the input $\vb*{x}^{(i)}$.

\subsection{Methodology}

Our goal is to obtain a function $f$ such that it produces significance values matching with the ranking information in the ground truth. Let $f: \mathbb{R}^d \rightarrow \mathbb{R}^{2K}$ be a trainable function parameterized by $\zeta$, where the output of the function is the predicted Gaussian distribution parameters $\hat{\mu}^{(i)} \in \mathbb{R}^K$ and $\hat{\sigma}^{2(i)} \in \mathbb{R}^K$ for the input $\textbf{\textit{x}}^{(i)} \in \mathbb{R}^d$ such that $f(\textbf{\textit{x}}^{(i)};\zeta) = [\hat{\mu}^{(i)} \; \hat{\sigma}^{2(i)}]$ and the predicted significance values $\hat{s}_j^{(i)} \sim \mathcal{N}\big(\hat{\mu}_j^{(i)}, \hat{\sigma^2}_j^{(i)}\big)$.

We adapt the optimization problem in Equation~\eqref{eqn:simple_mlr} as follows: instead of using a separate binary variable for the classification task, we model the significance values such that for a class ${y}_c$,  $\hat{s}_c^{(i)} \geq 0$ indicates a predicted positive, and $\hat{s}_c^{(i)} < 0$ indicates a predicted negative. The reformulated optimization problem reads:

\begin{equation}
    \label{eqn:gauss_mlr_prob}
    \max_{\zeta} \left[\prod_{c=1}^{K} T^{(i)}(\hat{s}_c^{(i)}, y_c) \smashoperator{\prod_{(y_u,y_v) \in \mathcal{B}^{(i)}}} P(\hat{s}^{(i)}_u \geq \hat{s}^{(i)}_v)\right],
\end{equation}
where
\begin{equation}
    T^{(i)}(\hat{s}_c^{(i)}, y_c) = P(\hat{s}_c^{(i)} \geq 0)^{\mathds{1}[y_c \in \mathcal{Y}^{(i)}]} P(\hat{s}_c^{(i)} < 0)^{\mathds{1}[y_c \notin \mathcal{Y}^{(i)}]},
\end{equation}

Letting $\hat{d}_{(u, v)}^{(i)} = \hat{s}^{(i)}_{u} - \hat{s}^{(i)}_{v}$, and using the above-mentioned definition: $\hat{d}_{(u, v)}^{(i)} \sim \mathcal{N}(\hat{\mu}^{(i)}_{u} - \hat{\mu}^{(i)}_{v}, \hat{\sigma}^{2(i)}_{u} + \hat{\sigma}^{2(i)}_{v})$, we can re-write Equation~\eqref{eqn:gauss_mlr_prob} as follows:
\begin{equation}
    \label{eqn:gauss_mlr_prob_with_diff}
    \max_{\zeta} \left[\prod_{c=1}^{K} T^{(i)}(\hat{s}_c^{(i)}, y_c) \smashoperator{\prod_{(y_u,y_v) \in \mathcal{B}^{(i)}}} P(\hat{d}_{(u, v)}^{(i)} \geq 0)\right],
\end{equation}

Considering a Gaussian random variable $z \sim \mathcal{N}(\mu, \sigma^2)$, probability of $z$ being positive can be written as a function $Q(\mu, \sigma)$, $$P(z > 0) = \frac{1}{2} \left[ 1 - \erf\left( -\frac{\mu}{\sigma \sqrt{2}} \right) \right] = Q(\mu, \sigma),$$ where $\erf(\cdot)$ denotes the Gaussian error function. This allows us to reformulate Equation~\eqref{eqn:gauss_mlr_prob_with_diff} as:

\begin{equation}
    \begin{split}
        &\max_{\zeta} \left[ \prod_{c=1}^{K} Q_r^{(i)}(\hat{\mu}_c^{(i)}, \hat{\sigma}_c^{(i)}, y_c)
        \smashoperator{\prod_{(y_u, y_v) \in \mathcal{B}^{(i)}}} Q(\hat{\mu}_{(u,v)}^{(i)}, \hat{\sigma}_{(u,v)}^{(i)}) \right], \\
        &\text{where }\\ &Q_r^{(i)}(\hat{\mu}_c^{(i)}, \hat{\sigma}_c^{(i)}, y_c) = 
        \begin{cases} 
            Q(\hat{\mu}_c^{(i)}, \hat{\sigma}_c^{(i)}) & \text{if } y_c \in \mathcal{Y}^{(i)}, \\
            1 - Q(\hat{\mu}_c^{(i)}, \hat{\sigma}_c^{(i)}) & \text{if } y_c \notin \mathcal{Y}^{(i)}.
        \end{cases}
    \end{split}
\end{equation}

Here $\hat{\mu}_{(u,v)}^{(i)} = \hat{\mu}^{(i)}_{u} - \hat{\mu}^{(i)}_{v}$ and $\hat{\sigma}_{(u,v)}^{(i)} = \sqrt{\hat{\sigma}^{2(i)}_{u} + \hat{\sigma}^{2(i)}_{v}}$. Applying negative log likelihood, we define our loss function in two parts as follows:

\begin{equation}
    \begin{split}
        L_c(\hat{\mu}^{(i)}, \hat{\sigma}^{(i)}, \mathcal{Y}^{(i)}) &= \sum_{c=1}^{K} -\log Q_r^{(i)}(\hat{\mu}_c^{(i)}, \hat{\sigma}_c^{(i)}, y_c),\\
        L_r(\hat{\mu}^{(i)}, \hat{\sigma}^{(i)}, \mathcal{B}^{(i)}) &= \smashoperator{\sum_{(y_u,y_v) \in \mathcal{B}^{(i)}}} -\log(Q(\hat{\mu}_{(u,v)}^{(i)}, \hat{\sigma}_{(u,v)}^{(i)})).
    \end{split}
\end{equation}

Summing up, the objective function of the UniMLR is given by:
\begin{equation}
    \label{eqn:gaussian_mlr_loss}
    \min_{\zeta} \dfrac{1}{N} \sum_{i=1}^N L_c(\hat{\mu}^{(i)}, \hat{\sigma}^{(i)}, \mathcal{Y}^{(i)}) + L_r(\hat{\mu}^{(i)}, \hat{\sigma}^{(i)}, \mathcal{B}^{(i)}).
\end{equation}

\begin{figure*}[ht]
    \centering
    \begin{subfigure}[b]{\linewidth}
        \hspace{0.10in}
        \includegraphics[width=0.48\linewidth]{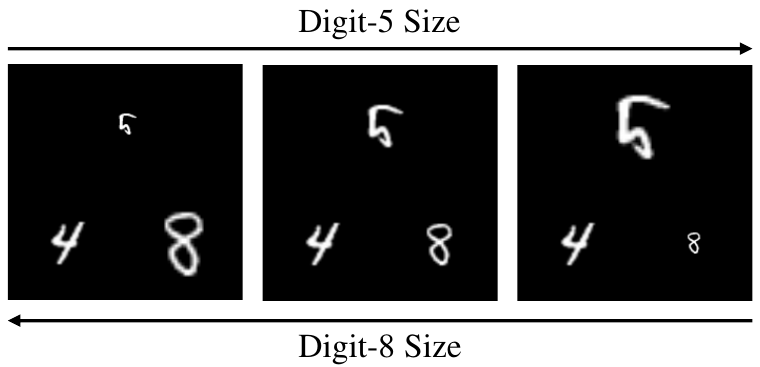}
        \includegraphics[width=0.48\linewidth]{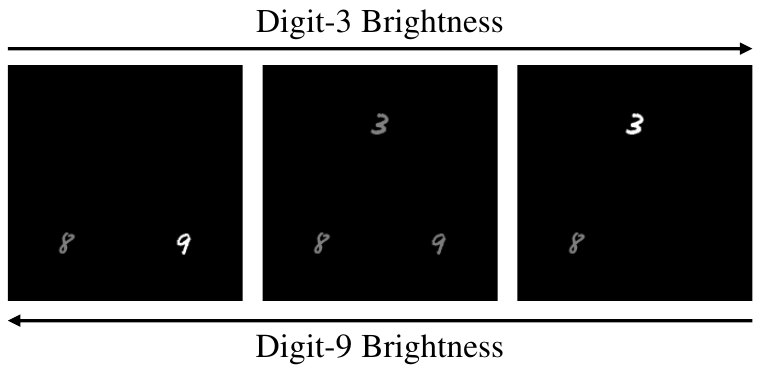}
    \label{fig:sequence}
    \end{subfigure}
    
    \begin{subfigure}[b]{\linewidth}
        \centering
    
        \setlength\tabcolsep{0.2pt}
        \begin{tabular}[b]{cccccc}
    
            \includegraphics[width=0.16\linewidth]{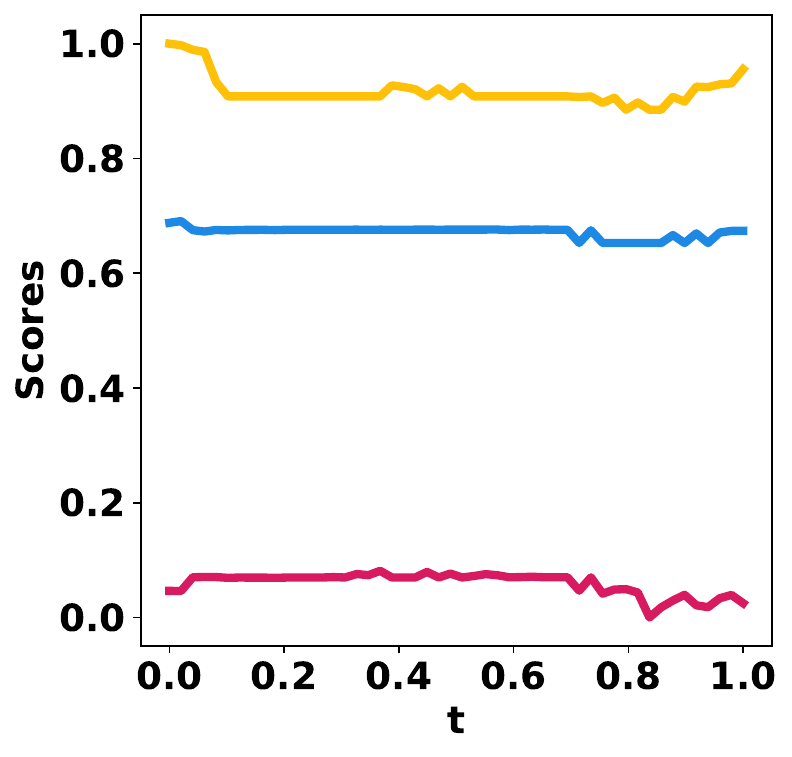} &
            \includegraphics[width=0.16\linewidth]{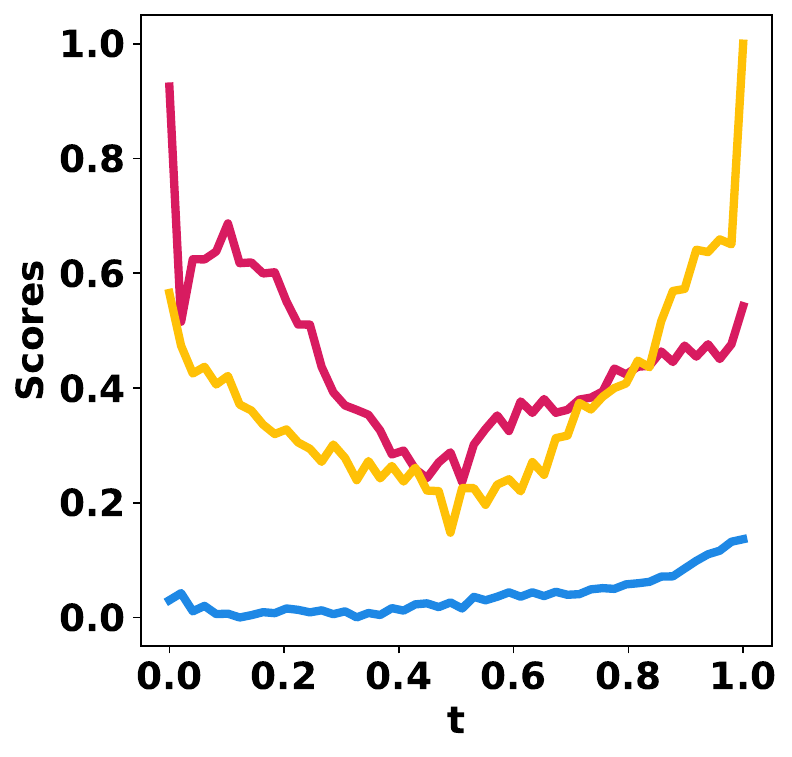} &
            \includegraphics[width=0.16\linewidth]{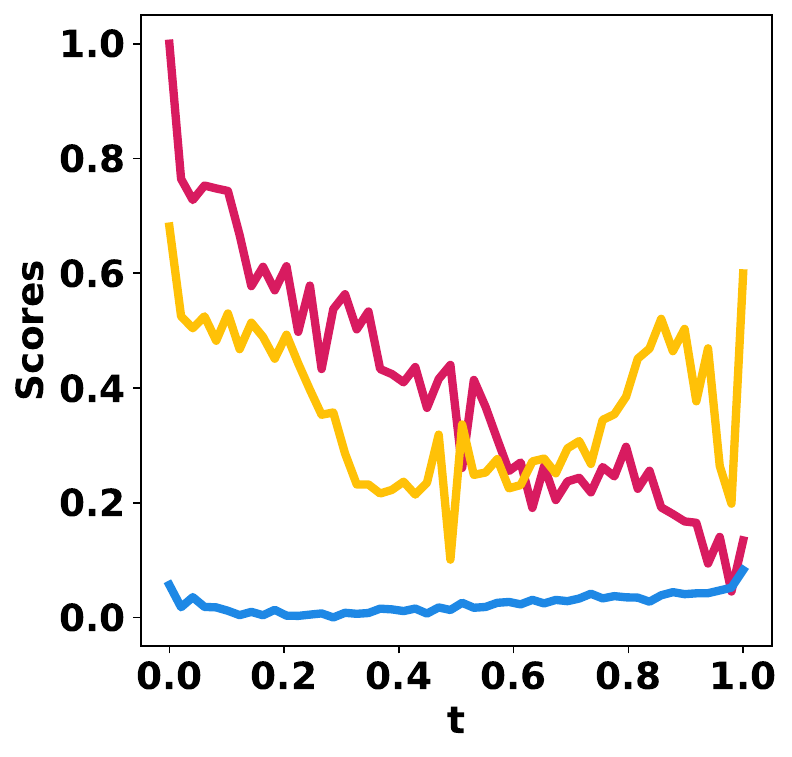} &
            \includegraphics[width=0.16\textwidth]{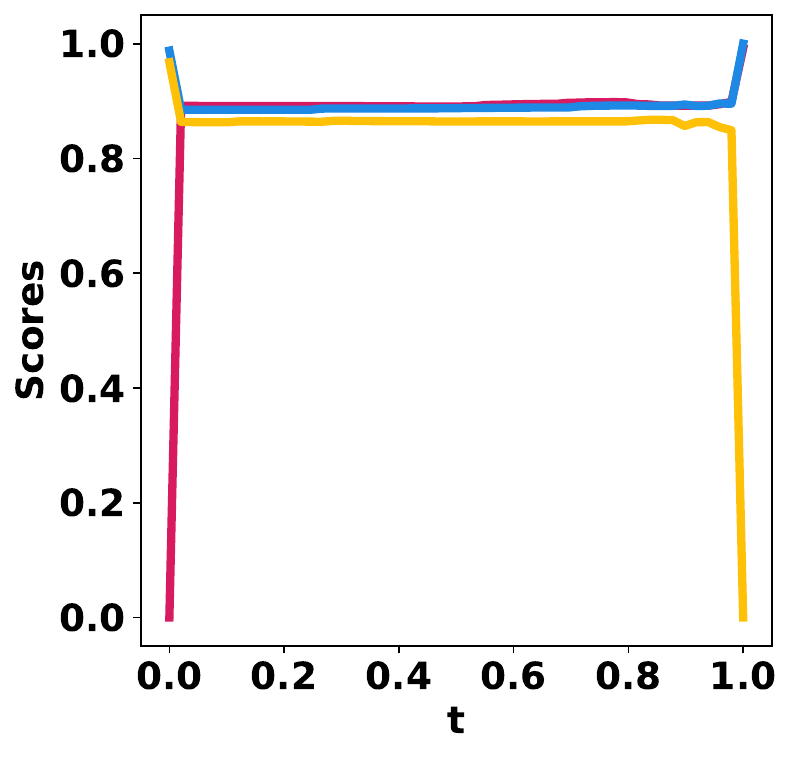} &
            \includegraphics[width=0.16\textwidth]{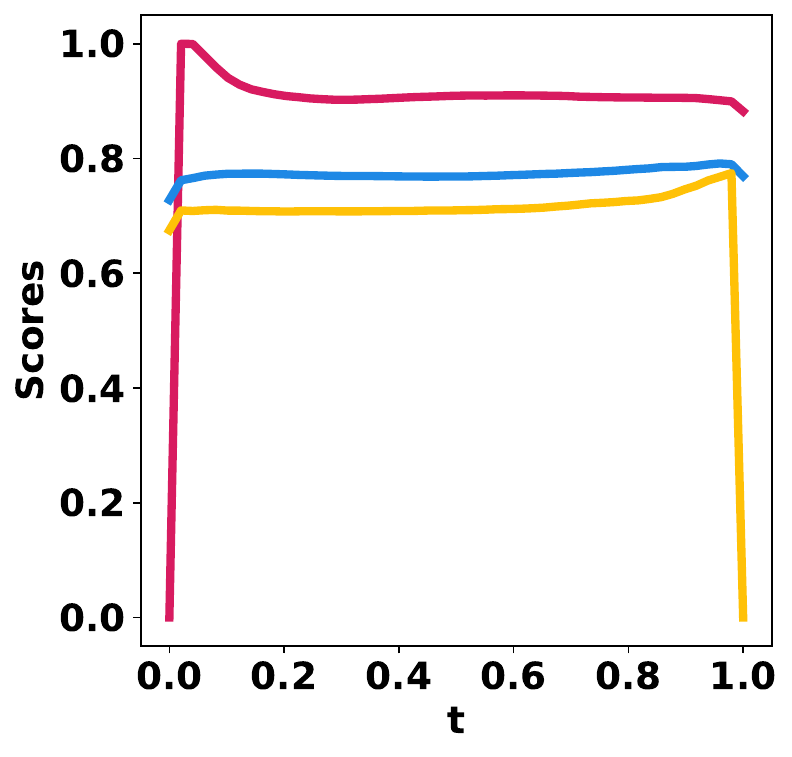} &
            \includegraphics[width=0.16\textwidth]{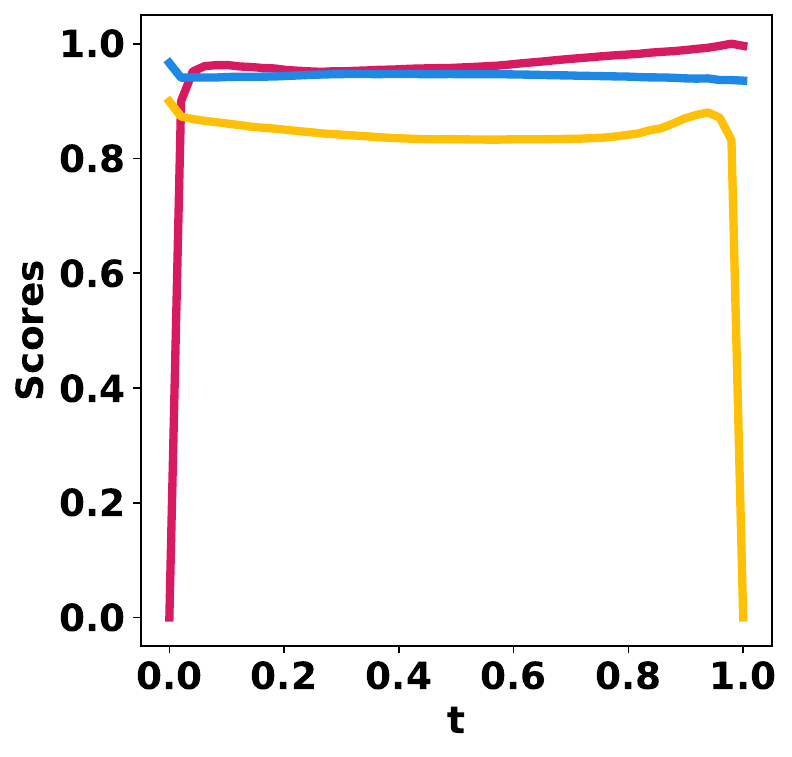} \\
            \hspace{0.12in} \small CRPC-Weak & \hspace{0.12in}  \small LSEP-Weak & \hspace{0.12in} \small UniMLR-Weak &   \hspace{0.12in} \small CRPC-Weak & \hspace{0.12in} \small LSEP-Weak & \hspace{0.12in} \small UniMLR-Weak
        
        \end{tabular}
    \end{subfigure}
    
    \begin{subfigure}[b]{\linewidth}
        \centering
        
        \setlength\tabcolsep{0.2pt}
        \begin{tabular}[b]{cccccc}
            &&&&&\\
            \includegraphics[width=0.16\linewidth]{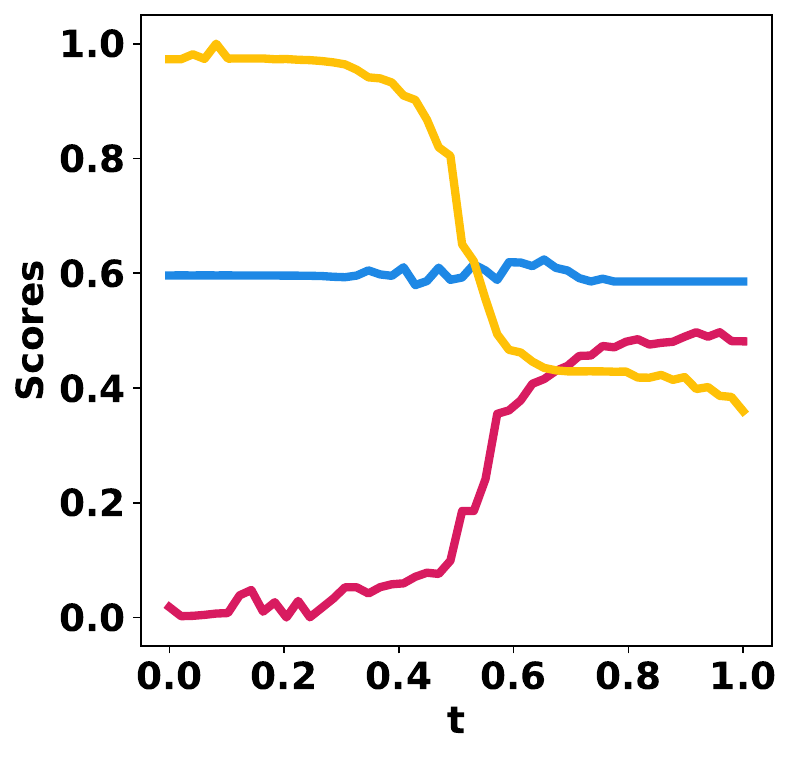} &
            \includegraphics[width=0.16\linewidth]{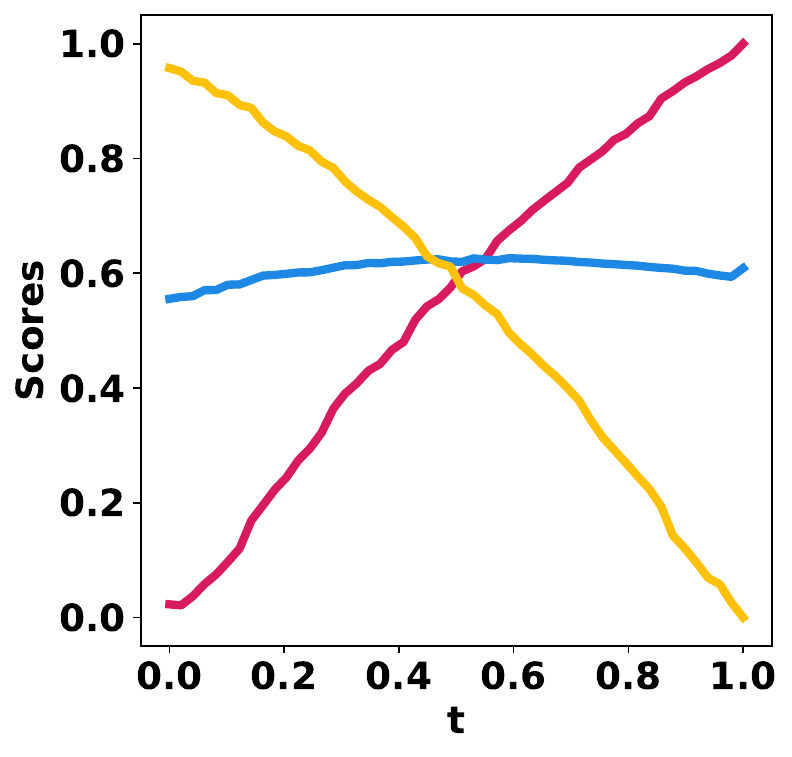} &
            \includegraphics[width=0.16\linewidth]{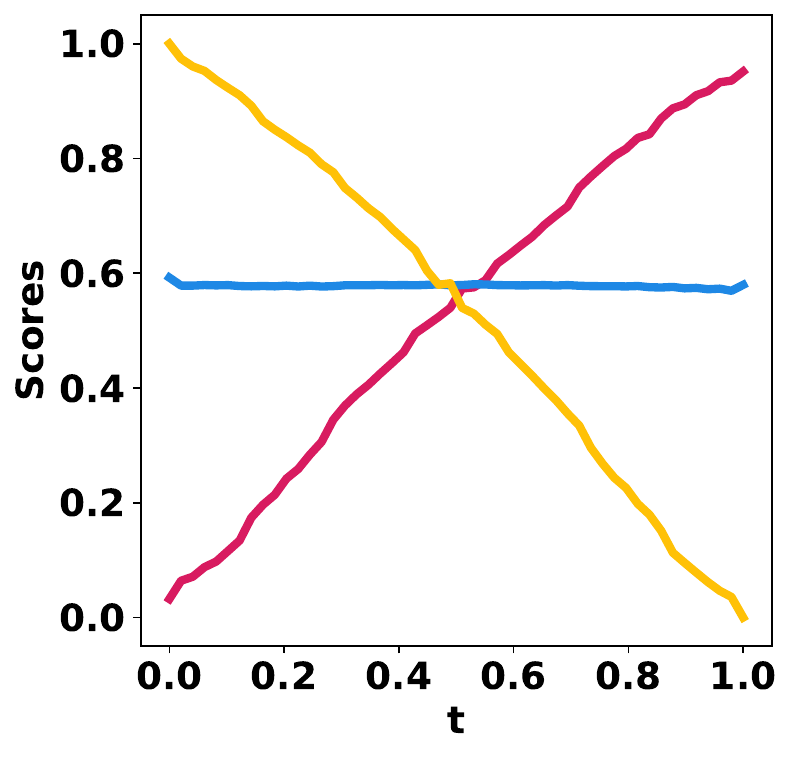} &
            \includegraphics[width=0.16\textwidth]{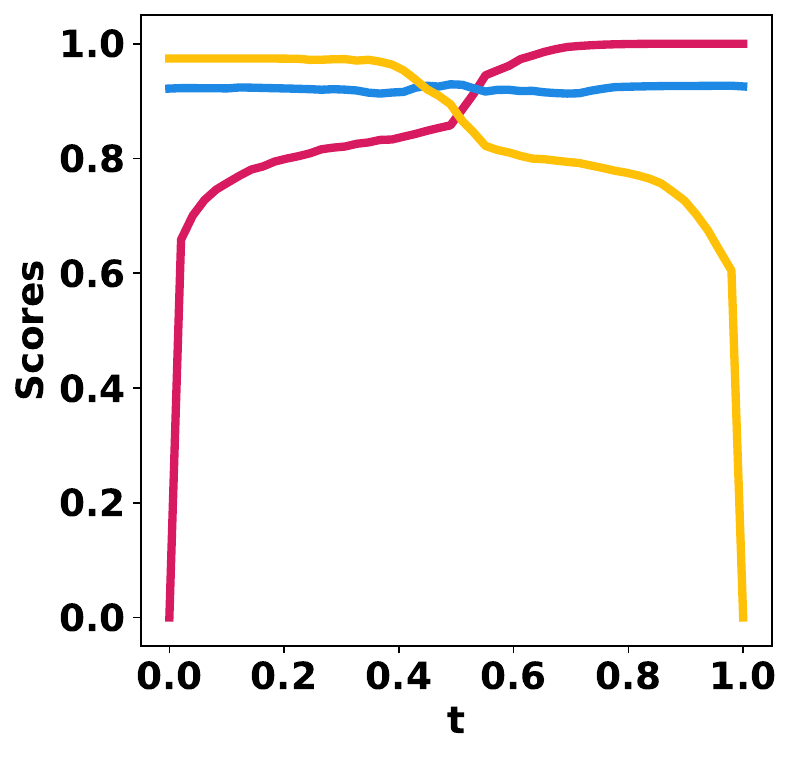} &
            \includegraphics[width=0.16\textwidth]{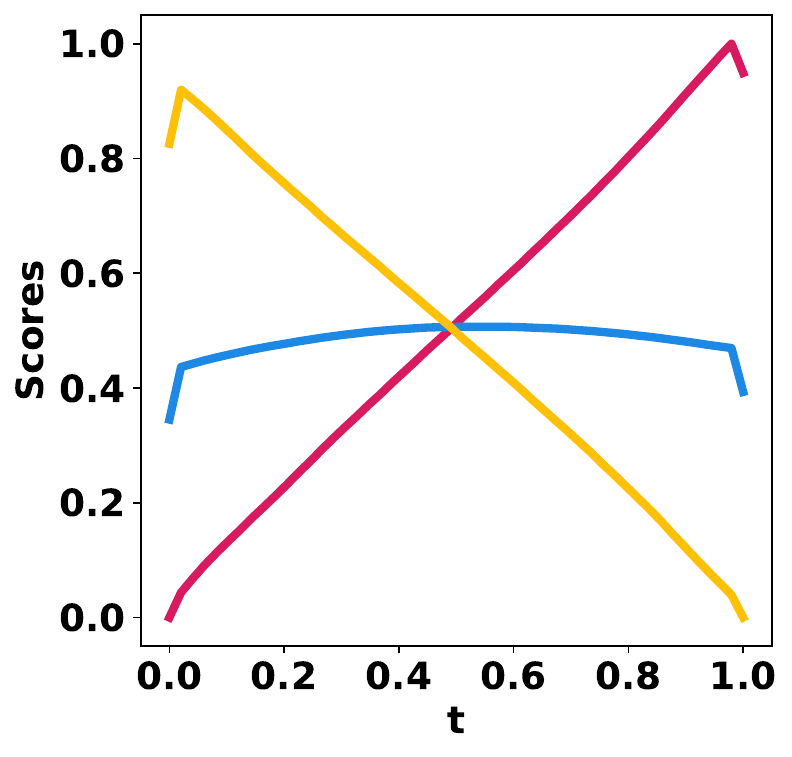} &
            \includegraphics[width=0.16\textwidth]{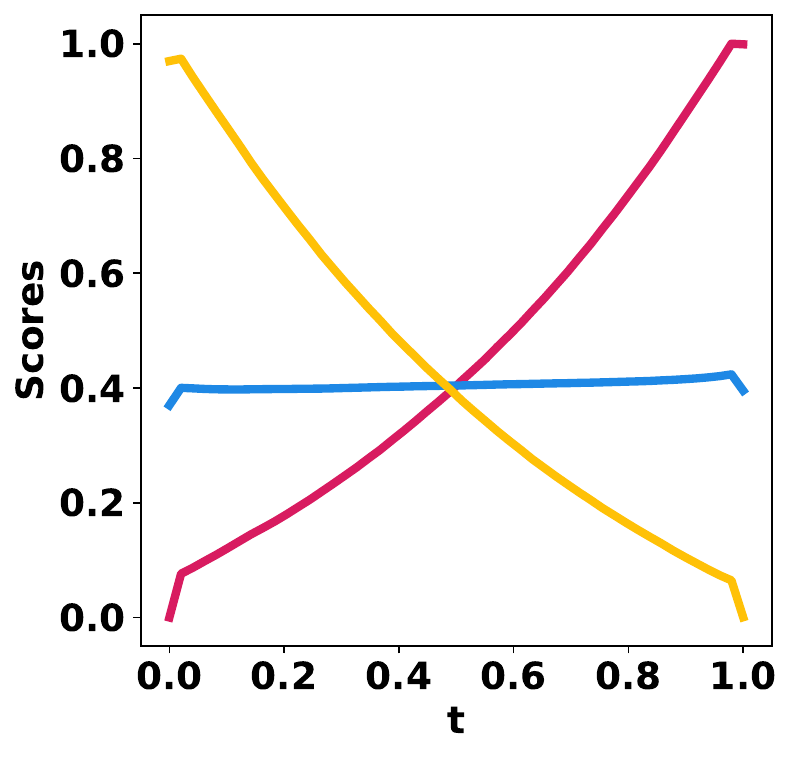} \\
             \hspace{0.12in} \small CRPC-Strong & \hspace{0.12in} \small LSEP-Strong & \hspace{0.12in} \small UniMLR-Strong &   \hspace{0.12in} \small CRPC-Strong & \hspace{0.12in} \small LSEP-Strong & \hspace{0.12in} \small UniMLR-Strong
        \end{tabular}
    \end{subfigure}
    \caption{Gradually changing significance effects in the sequences are shown at the top row, where the importance factor is the size of digits in top-left and brightness of digits in top-right. Lines demonstrate changes in scores of $\langle$\textbf{\textcolor{1st}{1st}}, \textbf{\textcolor{2nd}{2nd}}, \textbf{\textcolor{3rd}{3rd}}$\rangle$ digits, which are in the order of $\langle$5, 4, 8$\rangle$ in top-left and $\langle$3, 8, 9$\rangle$ in top-right. As UniMLR produces concurrently adjusting significance scores, it compares favorably over the baseline methods: CRPC and LSEP.}
    \label{fig:interpolation}
\end{figure*}

\textbf{Training and inference.} UniMLR provides a differentiable loss function (\ref{eqn:gaussian_mlr_loss}), thus in the context of our work we choose to use our objective to train neural networks using stochastic gradient descent. After training a network with any dataset $\mathcal{D}$, at inference we use $\hat{\mu}^{(i)}$ as the predicted ranking score for a given $\vb*{x}^{(i)}$.

\textbf{Learning implicit class significance.} For a large enough dataset, we claim that our predictions will be proportional to the real underlying significance values. We further support our claim with empirical studies in Section~\ref{sec:experiments}.

\section{Experiments}
\label{sec:experiments}

\subsection{Datasets}
\label{sec:datasets}
We conduct our experiments on three distinct datasets: the Natural Scene Images Database (NSID) \cite{InconsistentRankers}, the Architectural VDP Dataset (AVDP) \cite{DEMIR2021103826}, and the Ranked MNIST dataset, which we introduce separately in Section~\ref{sec:ranked_mnist}. These datasets have the common trait that they not only bipartite the labels into negatives and positives, but also provide how relevant each of these positive classes are to the instances. All datasets we use follow the notation provided in Section~\ref{sec:dataset_notation}. In the paper, we only provide experiments on Ranked MNIST Gray. 

\textbf{Natural Scene Images Database.} Natural scene images database \cite{InconsistentRankers} consists of 2000 images of natural scenes, for example: cloud, desert, mountain etc. The dataset has multiple labels per image, to convert them into a single label, we applied mean rank ordering \cite{Brinker2007CaseBasedMR} as a ranking aggregation method described in their paper. We randomly split images into sets with sizes 1600 and 400 to create our train and test sets accordingly.

\textbf{Architectural VDP Dataset.} Architectural Visual Design Principles (AVDP) \cite{DEMIR2021103826} dataset consists of 3654 train and 407 test images, where the labels are: asymmetric, color, crystallographic, flowing, isolation, progressive, regular, shape, symmetric. Images are associated with a maximum of 3 positive classes, and each positive class is ranked by its dominance over the other classes in representing the image. The authors of the paper can not provide the dataset publicly, and we obtained the dataset by asking from the authors.

\begin{figure*}[ht]
\centering
\vspace*{1cm}
    \begin{subfigure}[b]{1.0\linewidth}
        \centering
        \setlength\tabcolsep{0.2pt}
        \begin{tabular}[b]{ccccc}
            \includegraphics[width=0.14\linewidth]{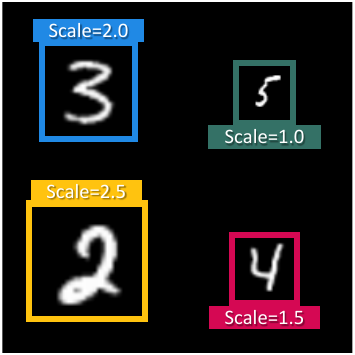} &
            \includegraphics[width=0.14\linewidth]{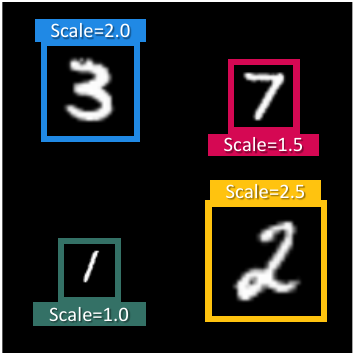} &
            \includegraphics[width=0.23\textwidth]{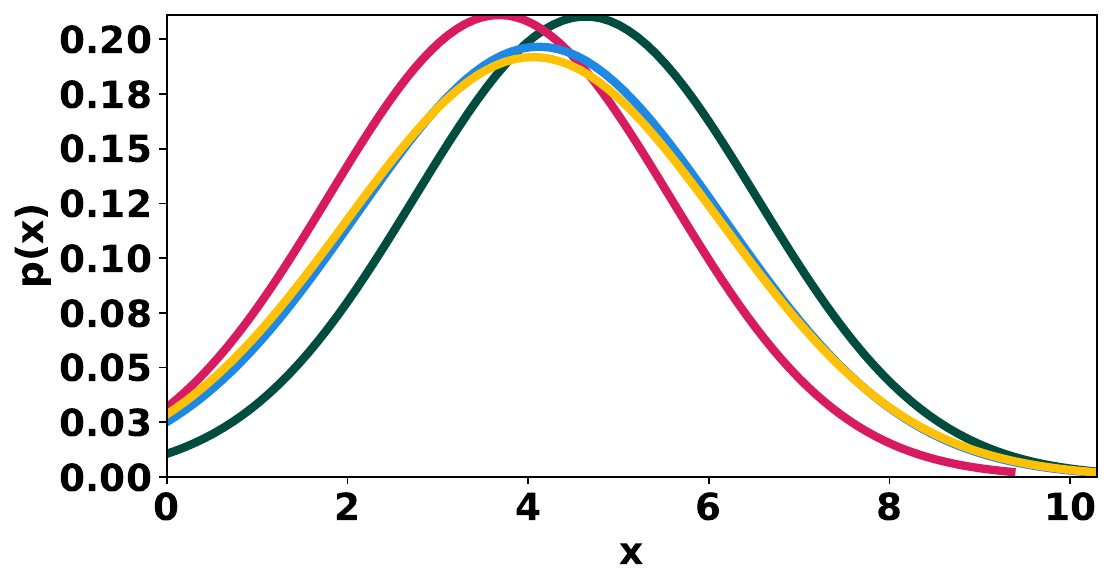} &
            \includegraphics[width=0.23\textwidth]{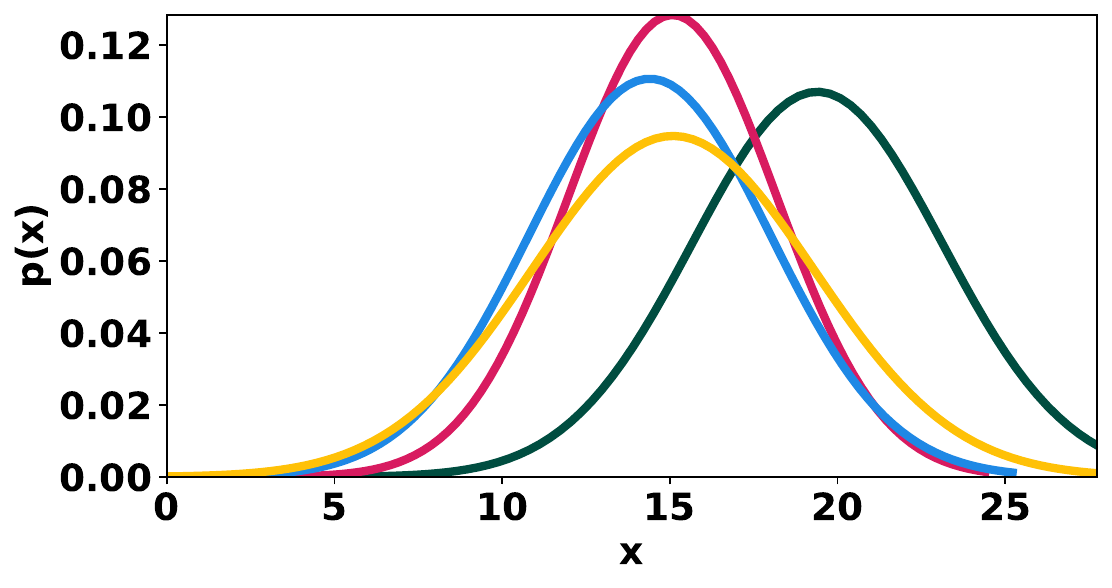} &
            \includegraphics[width=0.23\textwidth]{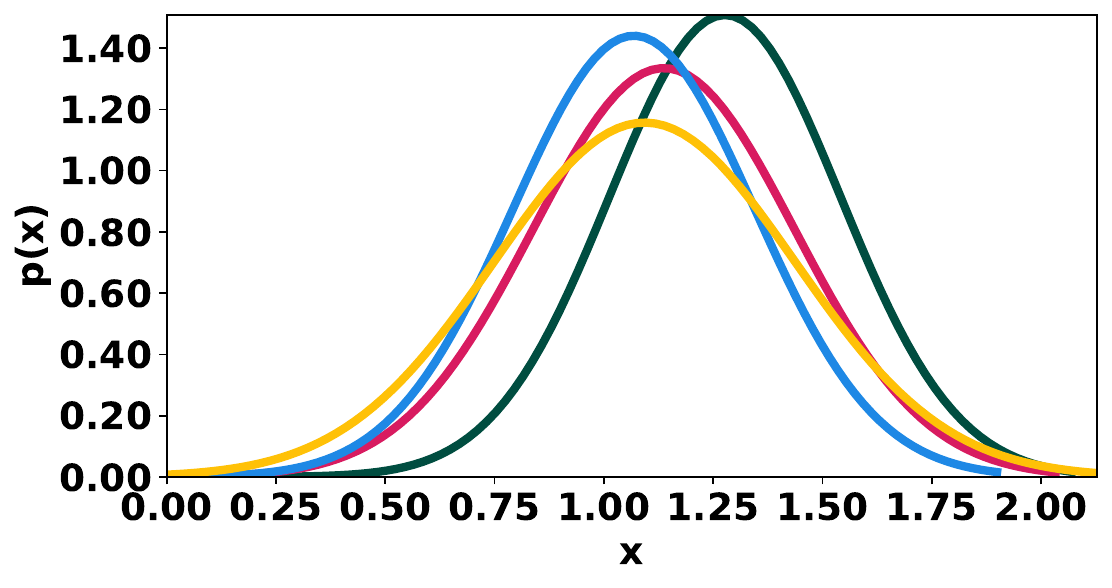}\\
            & &  \hspace{0.12in} CRPC-Weak & \hspace{0.12in}  LSEP-Weak & \hspace{0.12in}  UniMLR-Weak \\
        \end{tabular}
    \end{subfigure}
    
    \begin{subfigure}[b]{1.0\linewidth}
        \centering
        \setlength\tabcolsep{0.2pt}
        \centering
        \begin{tabular}[b]{ccccc}
            \includegraphics[width=0.14\linewidth]{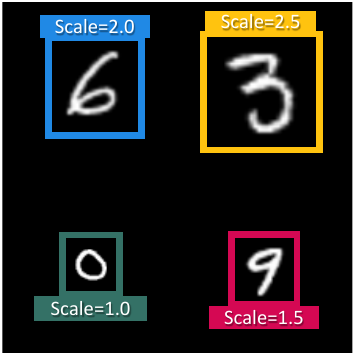} &
            \includegraphics[width=0.14\linewidth]{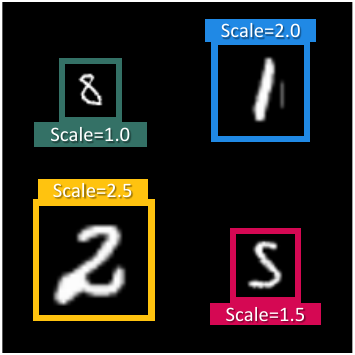} &
            \includegraphics[width=0.23\textwidth]{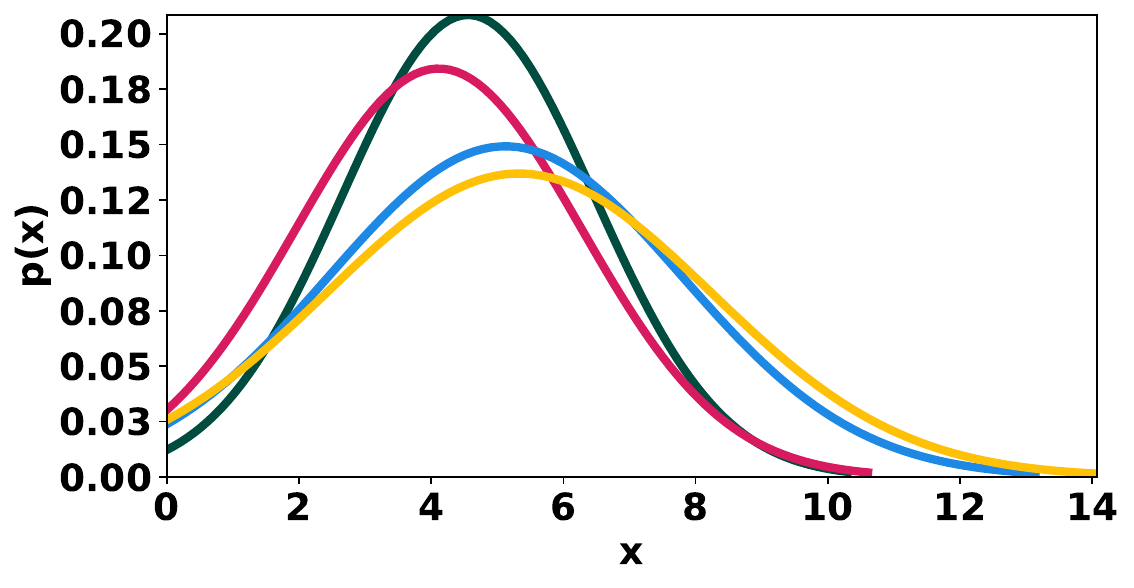} &
            \includegraphics[width=0.23\textwidth]{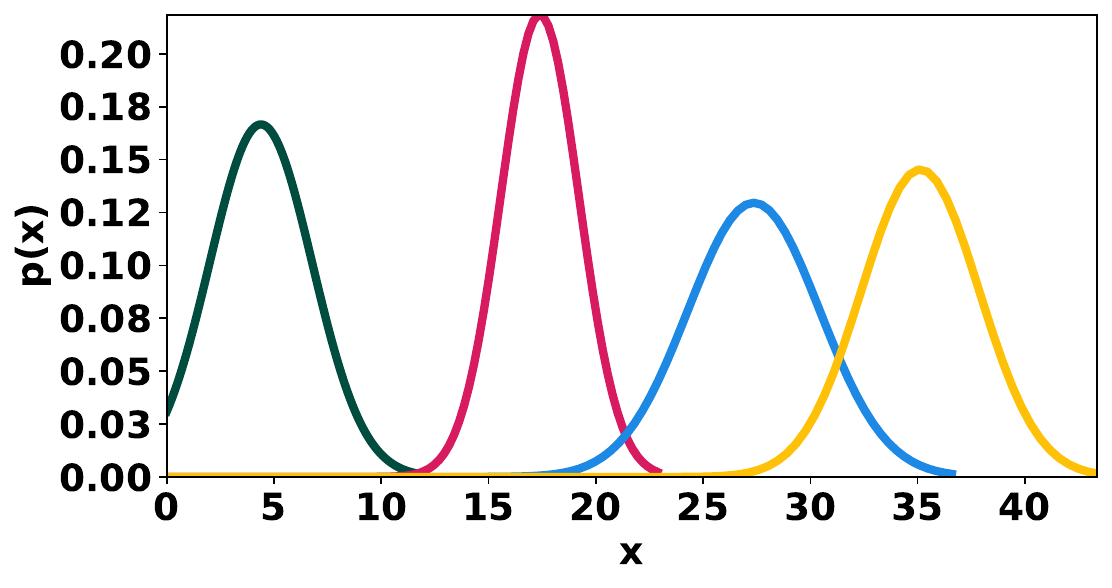} &
            \includegraphics[width=0.23\textwidth]{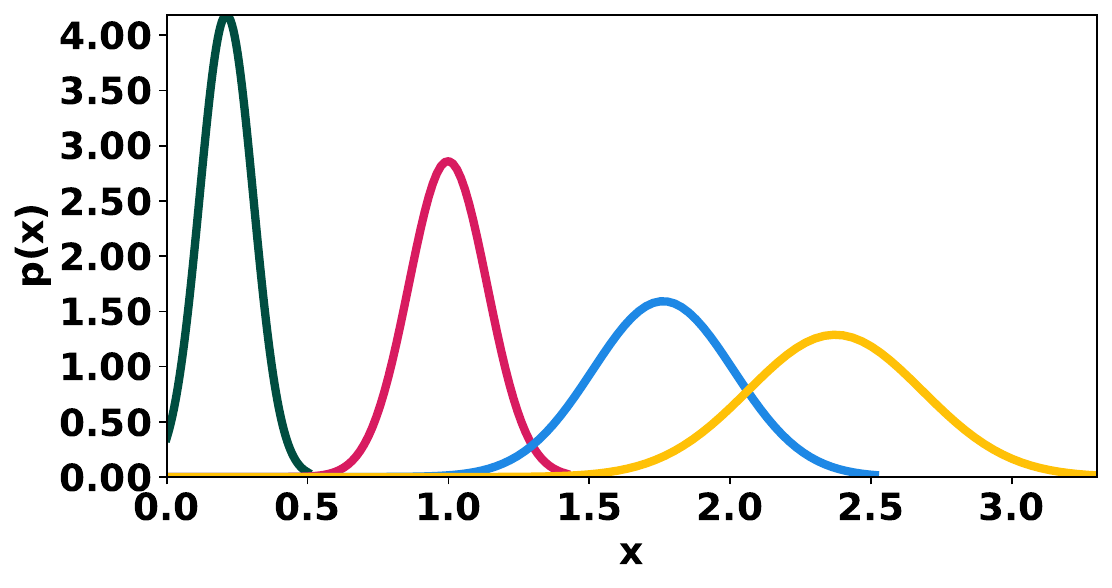}\\
            & & \hspace{0.12in} CRPC-Strong  & \hspace{0.12in} LSEP-Strong & \hspace{0.12in} UniMLR-Strong \\
        \end{tabular}
    \end{subfigure}
\caption{Annotated samples of 4-digit Ranked MNIST Gray-S dataset are shown in the first two columns from the left, where the scale factors of each digit are demonstrated by the bounding boxes for \textbf{\textcolor{4th}{Scale=1.0}}, \textbf{\textcolor{1st}{Scale=1.5}}, \textbf{\textcolor{2nd}{Scale=2.0}}, \textbf{\textcolor{3rd}{Scale=2.5}}. Resulting plots for each method when the set of scores for each significance value (scale) is fit to a Gaussian distribution are given in the other three columns. UniMLR captures proportional scores to significance values of each digit.} 
\label{fig:calibration}
\end{figure*}

\subsection{Ranked MNIST}
\label{sec:ranked_mnist}
Ranked MNIST is a family of datasets with two main branches named as Ranked MNIST Gray and Ranked MNIST Color, where the first is in grayscale and the latter has varying hue and saturation values for each digit. These datasets are generated by placing unique digits from the MNIST dataset \cite{deng2012mnist} on a 224x224 canvas, where the number of digits in a single image vary from 1 up to 10. Each branch has two importance factors that change: scale and brightness. According to these factors, we rank each positive digit such that for scale: the larger digits have greater ranks and for brightness: the brighter digits have greater ranks. For both Ranked MNIST Gray-S/B (Scale/Brightness) and Color-S/B datasets we have four different setups: changing scale, brightness, changing both and training on scales, changing both and training on brightness.

Ranked MNIST serves as a method for generating controllable, semi-synthetic MLR datasets with explicit rankings based on visual attributes. While designed for benchmarking, its broader utility is enabling controlled studies of label significance, something current MLR datasets lack.

\subsection{Baselines}

To evaluate our UniMLR method, we selected two pairwise baseline methods, namely: CRPC \cite{InconsistentRankers}, which is the calibrated \cite{calibrated_lr} version of RPC, and LSEP \cite{DBLP:journals/corr/LiSL17}. To our knowledge, UniMLR is the first multi-label ranking method which utilizes the positive class ranks, thus we aim \textit{to provide fairness in competition of the baseline algorithms} with our method. To that end, \textbf{we introduce CRPC-Strong} and \textbf{LSEP-Strong}, where we develop and change the existing baselines into \textit{Strong} versions that can process the positive class ranks by adding all $(y_u, y_v)$ pairs, where $y_u$ and $y_v$ are positive classes and $y_u \succ y_v$. Similarly, we call the methods which do not use the positive class relations as \textit{Weak} versions, where all positive labels equally have rank "1" and negatives rank "0". By introducing the weak/strong framework, we effectively introduce a new paradigm to the field and expand the scope of the prior work.

\subsection{Adjusting significance effects experiment}
\label{55}
To analyze the learned rank scores of UniMLR and baseline methods, we first conduct an experiment where we gradually adjusted the selected effects. We have two setups in the experiment: changing scale and changing brightness. For each setup, we generate a set of sequences $\mathcal{D}_a = \{\mathcal{S}_1, ..., \mathcal{S}_{50}\}$, where each sequence consists of gradually changing images, i.e. $\mathcal{S}_i = \langle \vb*{x}_{i}^{(1)}, ..., \vb*{x}_{i}^{(50)} \rangle$, here 50 is an arbitrarily chosen length for the sequence. Each sequence $S_i$ consists of three random MNIST digits, let us name them $y_i^{low}$, $y_i^{middle}$ and $y_i^{high}$, the starting image of the sequence $\vb*{x}_i^{(1)}$ has the corresponding significance values $s^{low}$, $s^{middle}$ and $s^{high}$. Iterating over the images of any sequence $\mathcal{S}_i$, the significance value for $y_i^{low}$ linearly changes from $s^{low}$ to $s^{high}$, for $y_i^{high}$ changes from $s^{high}$ to $s^{low}$, and $y_i^{middle}$ remains constant. In the top row of Figure \ref{fig:interpolation} how the images change for each setup are shown. For each method and setup, we obtain the rank scores of the images in $\mathcal{D}_a$ using the network trained with the corresponding method on the corresponding dataset Ranked MNIST Gray-S/B. The average value of each position over all sequences $\mathcal{S}_i$ are calculated for $y_i^{low}$, $y_i^{middle}$ and $y_i^{high}$, and we show how the predicted rank scores change in Figure \ref{fig:interpolation} for both strong and weak baselines, and UniMLR.

\subsection{Calibration experiment}
\label{56}
The calibration experiments are conducted to see how the scores for each baseline are distributed for different instances of the same significance values. We start by generating an image set $\mathcal{D}_C = \{\vb*{x}^{(1)}, ..., \vb*{x}^{(50)}\}$, where each $\vb*{x}^{(i)}$ consists of MNIST digits and is associated with a label set $\mathcal{Y}^{(i)} \subseteq Y$ where $|\mathcal{Y}^{(i)}| = 4$. Each positive class $y_j \in \mathcal{Y}^{(i)}$ in an image $\vb*{x}^{(i)}$ has a one-to-one mapping to $\{1.0, 1.5, 2.0, 2.5\}$ which defines their significance value, in this case the scale value. For each method in our experimental setup, we train a network on Ranked MNIST Gray-S, then we feed each image $\vb*{x}^{(i)}$ to the network to produce score vectors $\hat{\vb{s}}^{(i)}$. From each score vector $\hat{\vb{s}}^{(i)}$ we select the scores associated with each value in $\{1.0, 1.5, 2.0, 2.5\}$ then create a set of scores for each underlying significance value $\mathcal{S}_{1.0}$, $\mathcal{S}_{1.5}$, $\mathcal{S}_{2.0}$, $\mathcal{S}_{2.5}$. For each set we fit a Gaussian distribution to the values, the resulting plots for each method are given in Figure \ref{fig:calibration} with a visual expressing the experiment. Both LSEP and UniMLR extract an inherent calibration of significance scores, whereas in terms of their magnitude and variance, the order of distributions for each digit is best reflected by UniMLR. Another observation from our experiments is that, which is also exemplified in Figure \ref{fig:calibration}, UniMLR produces significance scores with larger variance for larger objects. This is due to naturally increased data variations owing to larger object extent. This positively distinct impact is not observed with other methods that entail intrinsic noise surpassing this effect.

\begin{table}[ht]
    \caption{Quantitative results on Ranked MNIST Gray, NSID and AVDP. Ranked MNIST S and B stands for changing scale or brightness of the digits, while (Mix) means both of the features are changing, but the ground truth indicates only one of the features. Bold-marked results show the best scores in Strong (S) baselines.}
    \centering
        \begin{tabular}{clcccc}
        \hline
         Dataset & Method & $\tau_b \uparrow $ & $ S \rho \uparrow$  & $\gamma \uparrow$ & F1 $\uparrow$ \\ \hline
          & CRPC (W) & 49.26 & 59.92 & 59.80 & 86.45 \\
          & LSEP (W) & 61.38 & 70.67 & 61.49 & \textbf{99.56} \\
         Ranked MNIST & UniMLR (W) & 62.52 & 71.86 & 62.62 & \textbf{99.54} \\
         Gray-S & CRPC (S)& 64.09 & 75.56 & 75.20 & 85.37 \\
          & LSEP (S)& 93.99 & 97.35 & \textbf{94.50} & 98.75 \\
          & UniMLR (S)&\textbf{94.23} & \textbf{97.41}  & 94.43 & 99.47 \\
         \hline
         
          & CRPC (W) & 51.36 & 61.45 & 59.27 & 89.44 \\
          & LSEP (W) & 59.45 & 68.67 & 59.77 & \textbf{99.12} \\
         Ranked MNIST & UniMLR (W) & 60.18 & 69.36 & 60.54 & \textbf{99.12} \\
         Gray-B & CRPC (S) & 61.71 & 73.62 & 74.70 & 81.73 \\
          & LSEP (S)&\textbf{93.62} & \textbf{97.01} & 94.46 & 98.21  \\
          & UniMLR (S) & 93.38 & 96.65 & \textbf{94.49} & 98.15\\
         \hline
         
          & CRPC (W) & 52.35 & 62.50 & 59.29 & 89.42 \\
          & LSEP (W) & 60.04 & 69.20 & 60.31 & 99.10 \\
          & UniMLR (W)&  60.00 & 69.23 & 60.24 & \textbf{99.15} \\
         NSID & CRPC (S) & 59.54 & 66.02 & 72.15 & 69.61\\
         & LSEP (S) & 72.57 & 76.57 & 80.58 & 80.12\\
         & UniMLR (S) & \textbf{75.44} & \textbf{78.66}  & \textbf{82.87} & 80.08 \\ \hline
         & CRPC (W) &  37.57 & 40.24 & 41.70 & 50.75\\
         & LSEP (W) & 39.79 & 41.93 & 44.54 & 54.66\\
         & UniMLR (W) & 40.29 & 42.69 & 41.12 & 54.31\\
         AVDP & CRPC (S) & 39.69 & 41.95 & 44.74 & 52.57\\
         & LSEP (S) &  40.54 & 42.60 & 42.55 & \textbf{55.86} \\
         & UniMLR (S) & \textbf{41.27} & \textbf{43.34} & \textbf{45.27} & 52.54\\ \hline
        \end{tabular}
        \label{tbl:quantitative}
\end{table}

\subsection{Quantitative results}

We provide quantitative results on both Ranked MNIST and real datasets using two set of metrics: ranking and classification. For ranking we use Kendall's Tau-b ($\tau_b$), Spearman's Rho ($S_\rho$) and Goodman and Kruskal's Gamma ($\gamma$). For classification we use F1 score. Table~\ref{tbl:quantitative} shows the results for each method trained on a different dataset. On Ranked MNIST Gray dataset, UniMLR slightly outperforms LSEP, and CRPC performs the worst amongst the three on all metrics. On real datasets, UniMLR outperforms the baselines for the ranking metrics, and produces comparable results to LSEP for the classification. It should be noted that both real datasets comprise inherent noise due to the labeling process being subjective. We observed that the weak methods have similar or better performance on classification metrics. We hypothesize that this is because strong methods must learn a full ranking, while weak methods only need to learn a bipartition, allowing the network’s entire capacity to focus on this simpler task.

\subsection{Variance experiment}
\label{sec:variance}
The variance experiments are conducted to observe whether our method distinguishes the small changes in importance factors and is able to assign ranks accordingly. The results of this experiment conducted on a dataset similar to Ranked MNIST Gray-S are provided in Table \ref{tab:small-variance}, with the scale sampled from $\mathcal{U}(1, 1.5)$ instead of  $\mathcal{U}(1, 3)$. Superior to other baselines, UniMLR-Strong captures the small differences in importance factors for both classification and ranking task.

\begin{table}[ht]
    \caption{Quantitative results of Variance Experiment on a variation of Ranked MNIST Gray-S where there are small changes in importance factors. The annotations of scores are the same with Table~\ref{tbl:quantitative}.}
    \centering
    \begin{tabular}{clcccc}
    \hline
    Dataset & Method & $\tau_b \uparrow$ & $S \rho \uparrow$ & $\gamma \uparrow$ & F1 $\uparrow$ \\
    \hline
    & CRPC (W)      & 50.57 & 60.94 & 59.81 & 88.05 \\
    & LSEP (W)      & 61.50 & 70.67 & 61.61 & 99.64 \\
    Ranked MNIST & UniMLR (W)    & 62.18 & 71.31 & 62.32 & 99.57 \\
    Gray-S & CRPC (S)      & 62.46 & 74.30 & 74.77 & 83.59 \\
    & LSEP (S)      & 92.10 & 96.50 & 92.61 & 98.79 \\
    & UniMLR (S)    & \textbf{92.65} & \textbf{96.77} & \textbf{92.86} & \textbf{99.52} \\
    \hline
    \end{tabular}
    \label{tab:small-variance}
\end{table}

\subsection{Error bars}
\label{sec:error_bars}
The error bar table for real datasets is given in Table \ref{tbl:error-bar-real}. It can be seen that Strong methods are consistently better compared to the weak methods. It should be noted that both of the real datasets consist of noisy and subjective labels which can affect the classification scores, while pairwise ranking explains the performance more accurately. Here UniMLR manages to outperform the rest of the methods on most of the ranking metrics and yields comparable scores for classification, due to the noisy nature of the datasets we can say the difference between UniMLR and LSEP for the classification does not strongly indicate that one is better while the other is not. We ran each of the training setup given in Table \ref{tbl:error-bar-real} for 5 times with random seeds, the scores on the table are the mean and standard deviation of the metrics for each random run.

\begin{table}[ht]
    \caption{Error bars for real datasets NSID and AVDP. Mean scores and standard deviations of each baseline after 5 runs are reported. The annotations of scores are the same with Table \ref{tbl:quantitative}.}

    \centering

        \begin{tabular}{lcccc}
        \multicolumn{5}{c}{Dataset: NSID} \\
        \hline
         Method & $\tau_b \uparrow $ & $ S \rho \uparrow$  & $\gamma \uparrow$ & F1 $\uparrow$ \\ \hline
          CRPC (W) & 57.68 $\pm$ 1.0 & 64.39 $\pm$ 1.0 & \multicolumn{1}{c}{70.43 $\pm$ 0.6 } & 67.97 $\pm$ 1.0 \\
          LSEP (W) & 72.8 $\pm$ 0.6 & 77.0 $\pm$ 0.6 & \multicolumn{1}{c}{80.3 $\pm$ 0.5} & 80.35 $\pm$ 0.3 \\
         UniMLR (W) &  {73.49 $\pm$ 0.9} & {77.84 $\pm$ 1.0} & \multicolumn{1}{c}{{80.96 $\pm$ 1.0}} & {80.63 $\pm$ 0.6} \\
          CRPC (S)& 59.34 $\pm$ 0.5 & 65.87 $\pm$ 0.4 & \multicolumn{1}{c}{72.47 $\pm$ 0.5} & 69.44 $\pm$ 0.4 \\
           LSEP (S)&  71.95 $\pm$ 1.8 & 75.94 $\pm$ 1.9 & \multicolumn{1}{c}{79.25 $\pm$ 1.8} & 80.17 $\pm$ 0.4 \\
          UniMLR (S)& \textbf{75.54 $\pm$ 0.4} & \textbf{78.86 $\pm$ 0.4} & \multicolumn{1}{c}{\textbf{82.89 $\pm$ 0.6}} & \textbf{80.21 $\pm$ 0.6}\\
         \hline
         \\
        \multicolumn{5}{c}{Dataset: AVDP} \\
        \hline
         Method & $\tau_b \uparrow $ & $ S \rho \uparrow$  & $\gamma \uparrow$ & F1 $\uparrow$ \\ \hline
         CRPC (W) & 38.71 $\pm$ 1.0& 41.29 $\pm$ 1.0 & \multicolumn{1}{c}{42.83 $\pm$ 1.9} & 51.8 $\pm$ 0.8 \\
         LSEP (W) &  39.21 $\pm$ 2.8 & 41.3 $\pm$ 3.0 & \multicolumn{1}{c}{41.93 $\pm$ 2.7} & {54.94 $\pm$ 1.2}\\
         UniMLR (W) & {41.13 $\pm$ 0.4} & {43.41 $\pm$ 0.4} & \multicolumn{1}{c}{{43.48 $\pm$ 1.8}} & 54.31 $\pm$ 0.5\\
        CRPC (S) & 39.91 $\pm$ 0.6 & 42.43 $\pm$ 0.6 & \multicolumn{1}{c}{44.71 $\pm$ 1.2} & 52.66 $\pm$ 0.3\\
         LSEP (S) &   40.95 $\pm$ 1.7 & 43.02 $\pm$ 1.8& \multicolumn{1}{c}{\textbf{44.98 $\pm$ 1.5}} & \textbf{55.42 $\pm$ 0.7} \\
         UniMLR (S) & \textbf{41.68 $\pm$ 1.6} & \textbf{43.84 $\pm$ 1.7} & \multicolumn{1}{c}{44.56 $\pm$ 2.8} & 54.12 $\pm$ 1.5\\ \hline
        \end{tabular}

        \label{tbl:error-bar-real}
\end{table}

\subsection{Extracted significance value experiment}
\label{sec:extracted}
\begin{figure*}[ht]
    \centering
    \includegraphics[width=0.9\textwidth]{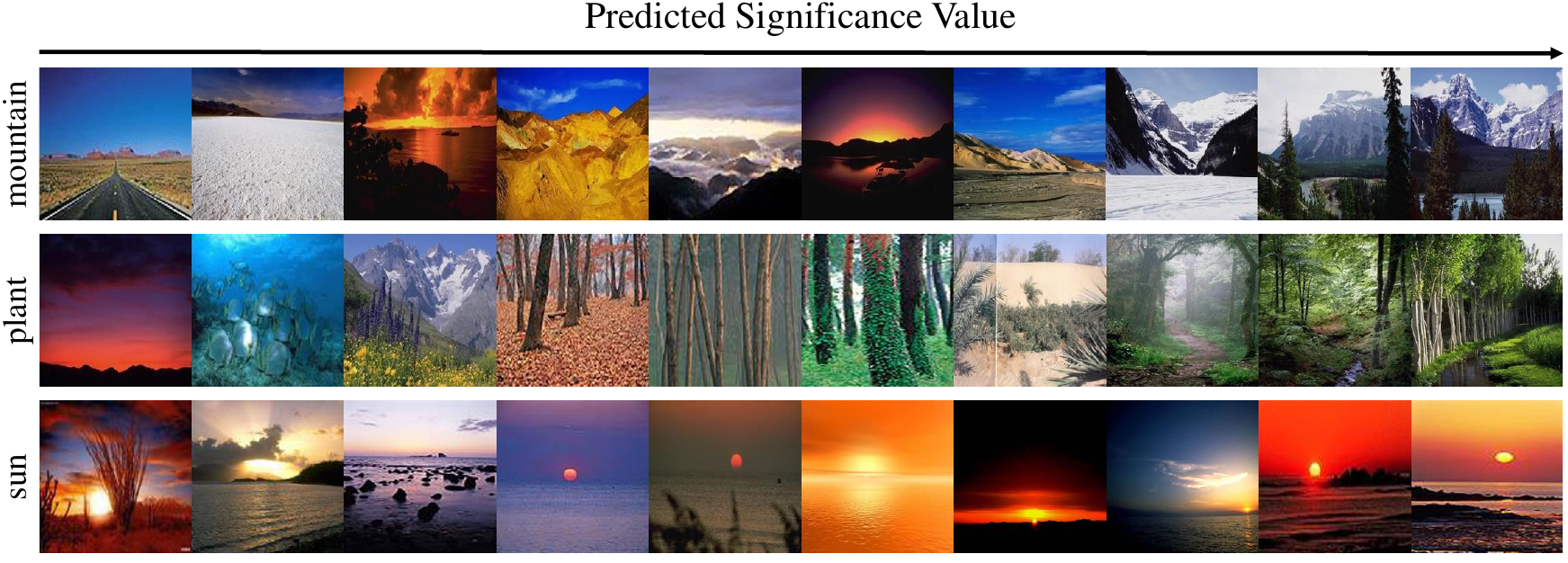} 
    \caption{Three sequences of images sampled from the test set of NSID, sorted in the order of predicted significance values for each class by UniMLR-Strong. The ordering of images demonstrates that UniMLR extracts class significance values that determine their rank proportionally to the dominant class in the image.}
    \label{fig:int-gmlr}
\end{figure*}

\begin{figure*}[ht]
    \centering
    \includegraphics[width=\textwidth]{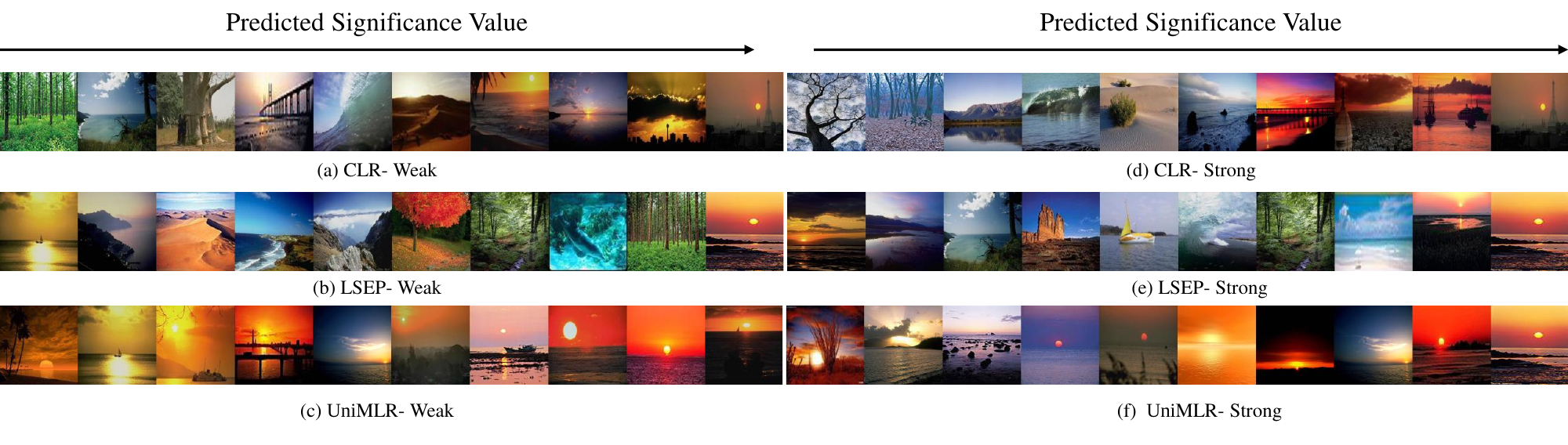} 
    \caption{Sun class benchmark for the extracted significance value experiment. The images are sorted by the significance value of the sun in that image, assigned by the corresponding methods. From left to right, the predicted significance of the Sun in the image increases.}
    \label{fig:sig_sun}
\end{figure*}

In this experiment, instead of creating gradually changing images to test if an MLR method produces consistent predictions with the underlying process, we do the opposite. Assuming the underlying process exists, we visualize which images in our test set would generate consistent scores with the process. We order the images according to the predicted significance value by our trained model for each class. Then, we select 10 images as checkpoints among the set of 400 test images, by choosing equidistant images in the interval. These image sequences, sorted as in Figure \ref{fig:int-gmlr}, demonstrate that as the predicted significance value for a class increases, the dominance of that class also increases. These results suggest that UniMLR extracts a proportional score to the underlying significance value for a class. 
In order to benchmark against other baselines, we provide the results of the Extracted Significance Value experiment for UniMLR and all baselines on NSID (sun) in Figure~\ref{fig:sig_sun}.

\section{Conclusion}
\label{sec:conclusion}
While previously studied weak multi-label ranking methods learn almost no useful information about the underlying significance values of the positive classes, the strong multi-label ranking paradigm of UniMLR yields remarkably calibrated significance values. UniMLR compares favorably to the competing baselines as demonstrated by the experiments, where the concurrent gradual changes in the scores for the changing effects in images as well as the constant scores for the static effects indicate that the underlying appearance and geometric characteristics pertinent to ranking are learned by UniMLR.

    
    
    

\paragraph{Broad impact.}

UniMLR encourages new ideas by providing a fresh perspective into the field of MLR. It introduces a set of datasets (Ranked MNISTs), which construct a controllable experimental environment for the new MLR paradigm.  Not only UniMLR shows the potential on learning more than a ranking between labels but also its output converges to a distribution that is proportional to the underlying significance value process of data characteristics. For a dataset where the labels of any instance are weighted by an unknown factor which controls their relevancy to the instance, UniMLR provides a way to extract these unknown factors by only using the ordering between the labels.   These findings we believe have potential in numerous  applications where ordering of factors are of value for instance in generation, design, or in decision making where alternative choices are typically pairwise ranked.
  
\paragraph{Limitations and ethical concerns. } 
 In the absence of strong MLR paradigm, it is an open question whether the MLR can reach or surpass the results provided by UniMLR.   The true potential of Strong MLR paradigm can be further appreciated with availability of more public synthetic and real-life datasets. 
 UniMLR does not impact the explainability and fairness of the decisions made by the underlying design choices, such as the architecture of the used neural nets in the pipeline. 
 UniMLR calls for further experimental and theoretical studies on learning calibrated significance values.






\bibliography{mybibfile}

\end{document}